\documentclass[letterpaper, 10 pt, conference]{ieeeconf}  

\IEEEoverridecommandlockouts                              

\usepackage{bm}
\usepackage[dvipsnames]{xcolor}
\usepackage{multicol}
\usepackage{amsmath}
\usepackage{graphicx}
\usepackage{amssymb}
\usepackage{float}
\usepackage{mathrsfs}
\usepackage{comment}
\usepackage{mathtools}
\usepackage{yfonts}
\usepackage{caption}
\usepackage{subcaption}
\usepackage{float}
\usepackage{algorithm}
\usepackage{algpseudocode}
\usepackage{xcolor}
\usepackage{cite}

\DeclareMathOperator*{\argmax}{arg\,max}

\newtheorem{assumption}{Assumption}

\usepackage{algorithm}
\usepackage{algpseudocode}

\definecolor{mycitecolor}{RGB}{71, 191, 38}
\definecolor{mylinkcolor}{RGB}{40, 115, 201}

\usepackage[bookmarks=true, colorlinks, citecolor=mycitecolor,linkcolor=mylinkcolor,urlcolor=mycitecolor]{hyperref}

\newcommand{\maulik}[1]{{\textcolor[rgb]{0,0,0}{#1}}}
\newcommand{\honghao}[1]{{\textcolor[rgb]{0,0,0}{#1}}}

\title{\LARGE \bf
Understanding and Imitating Human-Robot Motion with Restricted Visual Fields
}

\author{Maulik Bhatt$^{1\ast}$, HongHao Zhen$^{2\ast}$, Monroe Kennedy III$^{2}$, and Negar Mehr$^{1}$
\thanks{$^{1}$Department of Mechanical Engineering, University of California, Berkeley, 
        Berkeley, CA.
        Emails: \{maulikbhatt, negar\}@berkeley.edu
        }%
\thanks{$^{2}$Department of Mechanical Engineering,
        Stanford University, Stanford CA, USA.
        Emails: \{honghao\_zhen, monroek\}@stanford.edu
}
\thanks{$^\ast$Both authors contributed equally to this work.}
\thanks{This work is supported by the National Science Foundation, under grants ECCS-2438314 CAREER Award, CNS-2423130, and CCF-2423131.}
\thanks{Paper website \href{https://arm.stanford.edu/HRMotion}{https://arm.stanford.edu/HRMotion}}
 }

\begin{document}

\maketitle
\thispagestyle{empty}
\pagestyle{empty}

\begin{abstract}
When working around other agents such as humans, it is important to model their perception capabilities to predict and make sense of their behavior. In this work, we consider agents whose perception capabilities are determined by their limited field of view, viewing range, and the potential to miss objects within their viewing range. By considering the perception capabilities and observation model of agents independently from their motion policy, we show that we can better predict the agents' behavior; i.e., by reasoning about the perception capabilities of other agents, one can better make sense of their actions. We perform a user study where human operators navigate a cluttered scene while scanning the region for obstacles with a limited field of view and range. We show that by reasoning about the limited observation space of humans, a robot can better learn a human's strategy for navigating an environment and navigate with minimal collision with dynamic and static obstacles. We also show that this learned model helps it successfully navigate a physical hardware vehicle in real-time. Code available at \href{https://github.com/labicon/HRMotion-RestrictedView}{https://github.com/labicon/HRMotion-RestrictedView}.

\end{abstract}



\section{INTRODUCTION}


\vspace{-0.1cm}
As robots become more ubiquitous, they need to be able to operate in environments that are designed for humans as well as predict the behavior of humans and other agents in their proximity. In order to predict the motion of agents such as humans, the capabilities and limitations of a human must be understood by the robot. 
In this work, we argue that accurately modeling, predicting, and imitating human behavior requires us to account for the inherent limitations in human \emph{perception} and \emph{observation} capabilities.

\maulik{Evidence from cognitive science and behavioral studies supports this view. In sports, studies show that human gaze and field of view are strong indicators of attention and intended actions}~\cite{bertasius2017using, bertasius2018egocentric}. Even more generally, human intentions can be inferred from their gaze direction given their limited field of view \cite{umemoto2012search, wei2018and, belardinelli2023gaze,klingelschmitt2014combining}. These studies suggest that a human's field of view and point of view play a key role in understanding human intentions and predicting their actions.

\begin{figure}[t!]
    \centering
    \includegraphics[width=0.5\textwidth]{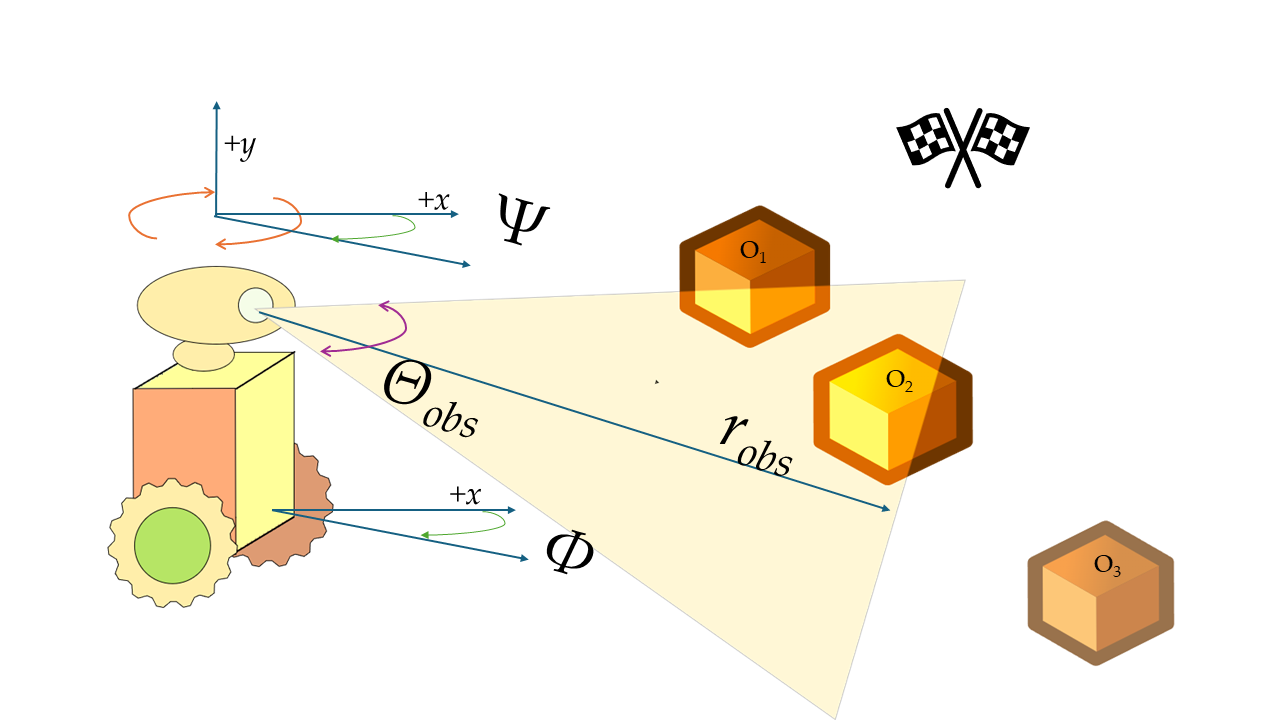}
    \caption{\small\textbf{Illustrative Example.} We present a robotic agent whose motion policy and observation ability are decoupled. The robot may move its base (orientation $\Phi$) and independently move its camera ($\Psi$). The observation of the camera has a limited range $r_{obs}$ and field of view $\Theta_{\text{obs}}$, and the likelihood of observing an object ($O_i)$ within the field of view. \maulik{Our goal is to accurately predict the motion of the agent navigating through the environment by observing its motion. In this work, we propose that the correct modeling of both the policy and observation model is necessary to predict the motion behavior of the agent accurately.}}
    \label{fig:running-example}
    \vspace{-0.5cm}
\end{figure}

Building on these findings, imitation learning seeks to replicate agent behavior, yet doing so effectively requires accounting for the demonstrator’s perceptual limitations. Current literature learns from demonstrations~\cite{schaal1996learning, argall2009survey, ravichandar2020recent} primarily through three approaches: inferring reward functions (Inverse Reinforcement Learning, IRL)~\cite{ng2000algorithms,ziebart2008maximum,mehr2023maximum}, directly learning and predicting the actions of the agent using Behavior Cloning Strategies (BC)~\cite{bain1995framework,ross2011reduction,daftry2017learning,torabi2018behavioral} or hybrid approaches where a discriminator must distinguish between artificial actions from a generator and actual demonstration actions \cite{ho2016generative,kang2018policy}. Diffusion models have recently emerged as state-of-the-art for imitation learning~\cite{sohl2015deep,ho2020denoising,chi2023diffusion, pearce2023imitating,10310116}. 
However, these models often neglect agents’ limited perception, which is critical for accurate imitation.
We argue that we need to account for the perception capabilities and observation space of the agents to better model, predict, and imitate agents such as humans.
Since actions depend on observations, robots that model limited perception of other agents, improve prediction accuracy.

\begin{figure*}
\hspace{0.1cm}
    \includegraphics[scale=0.27]{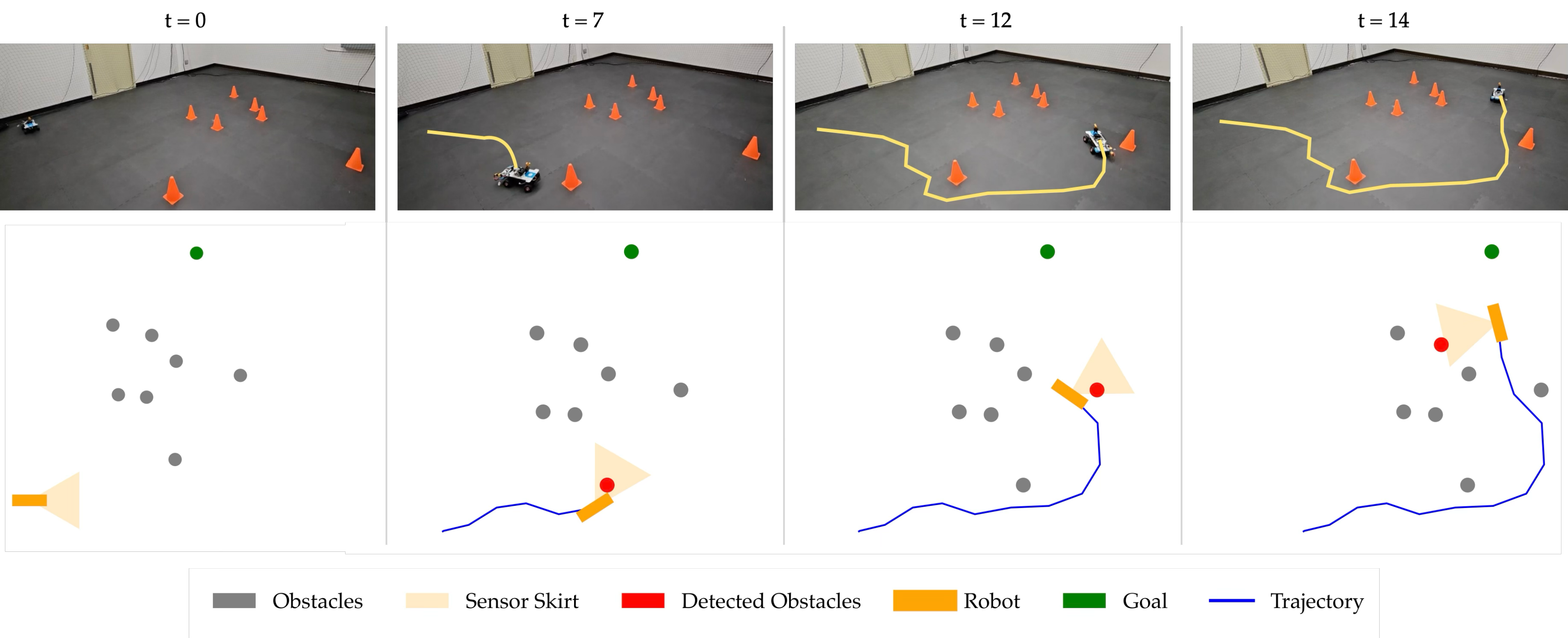}
    \vspace{-0.3cm}
        \caption{\small We employ the diffusion policy learned from human data on a car. 
        Like humans, the car changes its observation direction to look around, detect obstacles, and avoid them while moving toward the goal in real-time and successfully navigates the environment.}
    \label{fig:hardware-experiment}
    \vspace{-0.6cm}
\end{figure*}

We propose that to better model, predict, and learn agents' motion, it is necessary to correctly model both the \emph{observation model} of the agent and their policy. There are two key components that affect an agent's perception of an environment, which ultimately affect an agent's actions: (a) the field of view and range of the agent's visual observation, which determines where the agent is looking and what it can potentially perceive, and (b) the likelihood of successfully detecting objects within that observation region by the agent. In this paper, we examine the role of these two components and show that we can infer and estimate these parameters using demonstration trajectories of an agent. 

In this work, to better understand the role of an agent's observation strategy, we create a robotic agent with a limited field of view, view range, and likelihood of observing an obstacle, and use the agent's motion to infer these quantities. We consider three distinct scenarios, where each scenario involves one unknown: the parameters of the observation space, the probability of detecting obstacles within the agent's field of view, or the agent’s motion policy. In all scenarios, we show that by accounting for the agent’s inherent observation limitations, we can more accurately model and predict its behavior.

\maulik{It should be noted that while an end-to-end learning approach could, in theory, capture the complexities of human-like navigation, our modular approach offers several advantages in terms of interpretability, generalization, and robustness. By explicitly modeling the agent’s observation space, detection probability, and motion policy separately, we can better understand decision-making. This structure also enhances generalization, as the learned components can be adapted to different environments without requiring extensive retraining.  
This decomposition also isolates failures, making them easier to diagnose and mitigate compared to end-to-end learning, where errors can propagate unpredictably.
}

\maulik{In all scenarios, we show that by accounting for the agent’s inherent observation limitations, we can more accurately model and predict its behavior.} More specifically, we show that to infer the parameters of the observation space, one can use the cross-entropy optimization~\cite{rubinstein2004cross} to estimate the agent's observation parameters accurately. We then show that using Bayesian optimization~\cite{mockus2005bayesian}, the probability of object detection within an agent's field of view can be easily estimated. Finally, for inferring the policy of the agent, we show that the agent's policy can be accurately imitated using diffusion models~\cite{ho2020denoising}. We illustrate this by training a diffusion policy on human demonstrations of navigating an environment under a restricted field of view. We demonstrate our ability to predict and replicate human behavior in the presence of static and dynamic obstacles. Finally, we deploy the imitated policy to physical hardware, showcasing a robot navigating a cluttered environment and exhibiting human-like behavior by dynamically rotating its field of view to detect obstacles.

\section{METHODOLOGY}
\subsection{Problem Formulation}
Consider an agent navigating in an environment with obstacles. The obstacles could be either stationary or moving. We assume that the agent operates under partial observability as it can only perceive a limited portion of the environment at any time. This naturally aligns with a Partially Observable Markov Decision Process (POMDP) framework, where the agent maintains a belief over the true state of the environment based on incomplete observations.

We use $x_t \in \mathbb{R}^n$ to denote the state of the agent at time $t \in \mathbb{N}\cup\{0\}$, where is $\mathbb{N}$ is the set of natural numbers, and $n$ represents the agent's state space dimension. Let the control action of the agent at time $t$ be denoted by $u_t\in \mathbb{R}^m$ where $m$ is the dimension of the control space of the agent. We consider discrete deterministic dynamics for the agent to be denoted by the function $f:\mathbb{R}^n\times\mathbb{R}^m\rightarrow \mathbb{R}^n$, such that:
\begin{equation}\label{eq:dynamics}
    x_{t+1} =f(x_t,u_t).
\end{equation}
We assume the robot's dynamics are known and that the agent has limited observation capabilities. The agent's observation at time $t$ is denoted via $z_t \in \mathbb{R}^o$ where $o$ is the dimension of the observation space of the agent. We assume that the agent has an observation space that determines the area of the environment that the agent is looking towards. Furthermore, we assume that the agent has an inherent probability of observing something present in its observation area. With the introduction of this probability, we try to incorporate the scenarios in which humans sometimes miss objects even when they are present in their view. The combined observation space and probability of observation detection determine the agent's observation at each time, given the agent's current state. 

Let the observation space parameters that determine the agent's observation area be denoted by $\theta \in \Theta$ where $\Theta$ is the space of observation space parameters. Also, let the probability of observing a particular observation be denoted by $p_{obs} \in [0,1]$. These components define the agent's POMDP observation model, which we represent as
\begin{equation}
    P(z_t | x_t ; \theta, p_{obs}) = O(z_t|x_t;\theta, p_{obs}),
\end{equation}
where $O:\mathbb{R}^o\times\mathbb{R}^n \rightarrow [0,1]$ is the observation function parameterized by parameters $\theta$ and $p_{obs}$.

At each time $t$, based on the observation $z_t$, the agent takes a control action $u_t$. The probability of taking an action at each time $t$ by the agent is captured through a stochastic policy, which we denote via $\pi:\mathbb{R}^m\times\mathbb{R}^o\rightarrow[0,1]$,
\begin{equation}
    P(u_t|z_t;\theta_\pi) := \pi(u_t|z_t;\theta_\pi),
\end{equation}
where $\theta_\pi \in \Theta_\pi$ is the parameter that determines the policy for the agent. Here, $\Theta_\pi$ is the space of policy parameters.

We use $\tau$ to denote a trajectory of the agent as $\tau = \{x_0,u_0,x_1,u_1,\ldots,x_{T-1},u_{T-1},x_T\}$, where $T$ is the length of the trajectory. In this work, we assume that we have access to a set of $N$ trajectories, $\mathcal{D} = \{\tau_i\}_{i=1}^N$ exhibited by the agent in an environment with static or potentially moving obstacles. \textcolor{black}{As explained earlier, our main aim is to better model, predict, and learn the agent's behavior by observing its behavior in the environment. We do this by using a modular approach where we aim to separately estimate the agent's observation space parameter $\theta$ and policy parameter $\theta_\pi$ based on the trajectory data to mimic agent behaviors.}

\textbf{Running Example:} To illustrate this idea, consider an agent with discrete dynamics~\eqref{eq:dynamics} navigating through an environment toward its desired goal (Fig.~\ref{fig:running-example}). The agent perceives the scene with a visual sensor parameterized by the range of vision ($r_{obs}$), and the field of view ($\theta_{obs}$). Obstacles in the field of view are not guaranteed to be noticed (due to lack of visual features), which is captured by the detection probability $p_{obs}$. \maulik{Furthermore, the orientation in which robot looking, i.e. \emph{the orientation of the visual sensor} is denoted via $\psi_{obs}$. Note that the motion of the agent in any environment is dependent on the controls inputs chosen by the robot. Also, the agent can choose the orientation of the visual sensor to look in a particular direction.} Therefore, we assume that the motion of the agent is governed by a \maulik{joint} policy $\pi(\cdot;\theta_\pi)$, which controls a) the motion of the robot in the environment, and b) \emph{the orientation of the visual sensor} denoted by $\psi_{obs}$. With these constraints, the agent follows a POMDP-based navigation strategy, making decisions under uncertainty to minimize collisions while reaching the goal.

Given this example, we look at three key problems: \textbf{1)} estimating the agent's observation space parameters, \maulik{($r_{obs},\theta_{obs}$)}, \textbf{2)} calculating the probability of detecting an obstacle within the field of view, \maulik{$p_{obs}$}, and \textbf{3)} estimating the agent's motion policy for both displacement and sensor motion, \maulik{$\pi(\cdot;\theta_\pi)$} from a set of agent behavior demonstrations.


\maulik{To analyze the influence of each component independently, we adopt a modular approach: we estimate each unknown parameter while assuming the other two are known. This structure allows us to isolate the effect of each component on the agent's behavior and evaluate the accuracy of each estimation method.}

\subsection{Estimating the Observation Space}

First, we discuss how we can estimate the parameters of the observation space of an agent $\theta$ given a set of demonstrations of agent trajectories $\mathcal{D}$. We solve the problem through a maximum likelihood estimation approach. We make the following key assumption. 
\begin{assumption}
    We assume that the probability of observing an obstacle $p_{obs}$ within an observation space and the agent's policy parameters $\theta_\pi$ are known.
\end{assumption}

Given the knowledge of policy parameters for the agent and the probability of observing an obstacle, for a given initial state $x_0$ and deterministic dynamics~\eqref{eq:dynamics}, we can compute the probability of observing a trajectory $\tau$ under a specific value of $\theta$ as
\begin{align}\label{eq: prob of traj}
    & P(\tau;\theta) = P(x_0)\Pi_{t=0}^{T-1}P(x_{t+1}|x_t,u_t)\int P(z_t|x_t)P(u_t|z_t) d z_{t} \nonumber \\
    & = \Pi_{t=0}^{T-1}\int O(z_t|x_t;\theta,p_{obs})\pi(u_t|z_t(\theta);\theta_\pi) d z_{t}.
\end{align}


From now on, we omit denoting that integration is over $z_t$ for the sake of brevity. From \eqref{eq: prob of traj}, we can compute the probability of the whole dataset $\mathcal{D}$ as $P(\mathcal{D};\theta) = \sum_{i}^NP(\tau_i;\theta).$
We can alternatively maximize the log-likelihood of the trajectories: 
\begin{align}\label{eq:negative-log}
    \theta^* & = \argmax_{\theta \in \Theta}\sum_{i=1}^N\sum_{t=0}^{T-1}\log(\int O(z_t,x_t;\theta,p_{obs})\pi(u_t,z_t;\theta_\pi)).
\end{align}
\vspace{-0.4cm}

Solving \eqref{eq:negative-log} requires taking the gradient of the \maulik{objective} with respect to $\theta$, which would result in having to take the gradient of policy $P(u_t|z_t)$ with respect to $\theta$ where $z_t$ depends on $\theta$. 
\maulik{This is difficult due to its nested probabilistic dependencies. Furthermore, the presence of an integral over $z_t$ makes analytically computing the gradient of the objective intractable.}
Therefore, we resort to black-box optimization techniques such as Bayesian optimization \cite{mockus2005bayesian} and the cross-entropy method \cite{rubinstein2004cross}, which find the optimal solutions based on evaluations of the function without requiring us to compute the gradient of the objective. Bayesian optimization is a probabilistic model-based optimization technique used to optimize objective functions, which are expensive to evaluate. The Cross-Entropy Method is another optimization technique that iteratively samples random parameters, evaluates their performance, and refines the parameter distribution toward the optimal region~\cite{rubinstein2004cross}. \maulik{To estimate the observation space, we use the Cross-Entropy method.
}


In the following section, we provide an approach to estimate the probability of observation $p_{obs}$ of the agent.

\subsection{Estimating the Probability of Observation}
In order to estimate $p_{obs}$ given a set of demonstration trajectories $\mathcal{D}$, we make the following assumption
\begin{assumption}
    We assume the observation space parameters $\theta$ and the agent's policy parameters, $\theta_\pi$, are known.
\end{assumption}

Under this assumption, we can compute the probability of a trajectory given a value of $p_{obs}$ and find $p_{obs}$ that maximizes the likelihood. Estimating $p_{obs}$ can be cast as the following maximum likelihood estimation problem:
\begin{align}\label{eq:negative-log-2}
    & p_{obs}^* = \argmax_{p_{obs} \in [0,1]} \sum_{i=1}^N\sum_{t=0}^{T-1}\log(\int O(z_t|x_t;\theta,p_{obs})\pi(u_t|z_t;\theta_\pi)).
\end{align}
The problem in \eqref{eq:negative-log-2} is similar to that of \eqref{eq:negative-log}. Therefore, we can use the same black-box function optimization techniques. Since we need to estimate only one parameter $p_{obs}$, we employ Bayesian optimization.

\subsection{Estimating the Policy of the Agent}

In order to infer and learn the policy of the agent given the dataset $\mathcal{D}$, we make the following key assumption:
\begin{assumption}
    We assume that the observation space parameters $\theta$ are known.
\end{assumption}
Under this assumption, the problem of estimating $\theta_\pi$ can be treated as a behavior cloning problem, i.e., we aim to learn the policy $\pi(\cdot;\theta_\pi)$ parameterized by $\theta_\pi$ such that it imitates the provided data set $\mathcal{D}$ in the best possible way. \maulik{Unlike prior BC approaches that assume full observability, our method conditions the policy on what the agent can perceive, explicitly incorporating limited field of view and probabilistic detection. This leads to a more realistic imitation of the agent behavior.} If the policy is a neural network, then $\theta_\pi$ can be considered to be the neural network parameters. State of the art in behavior cloning is diffusion policies \cite{chi2023diffusion}. These methods model policy generation as a diffusion process, where an initial noisy signal or random trajectory is progressively refined toward optimal actions through iterative steps.  As such, we adopt diffusion policies for mimicking the agent's policy. We formulate policies such as Denoising Diffusion Probabilistic Models (DDPMs) conditioned on the state and observation of the agent. The detailed implementation for this method can be found in \cite{chi2023diffusion}.

In the following section, we showcase the capabilities of our method through extensive numerical simulations.

\section{Numerical Simulations}
In this section, we present numerical simulations to showcase the importance of considering agents' observation and perception capabilities. We consider synthetic settings where we have access to the ground-truth parameters, which enable a comprehensive comparison and evaluation of our method. Then, in the next section, we consider human data and perform hardware experiments that leverage actual human data. We consider a unicycle agent with discrete dynamics:
\begin{align}\label{unicycle}
    p_{t+1} & = p_t + h\cdot v_k\cos{\phi^i_t}, \;
    q_{t+1} = q_t + h\cdot v_t\sin{\phi^i_t} \nonumber \\
    \phi_{t+1} & = \phi_t + h\cdot\omega_t,
\end{align}
where $h$ is the duration of the time step, $p_t$ and $q_t$ are $x$ and $y$ coordinates of the positions in the 2D plane, $\phi$ is the heading angle from positive x-axis, $v_t$ is the forward velocity, and $\omega_t$ is the angular velocity of the agent at time $t$. 

\begin{figure}
    \includegraphics[width=1\linewidth]{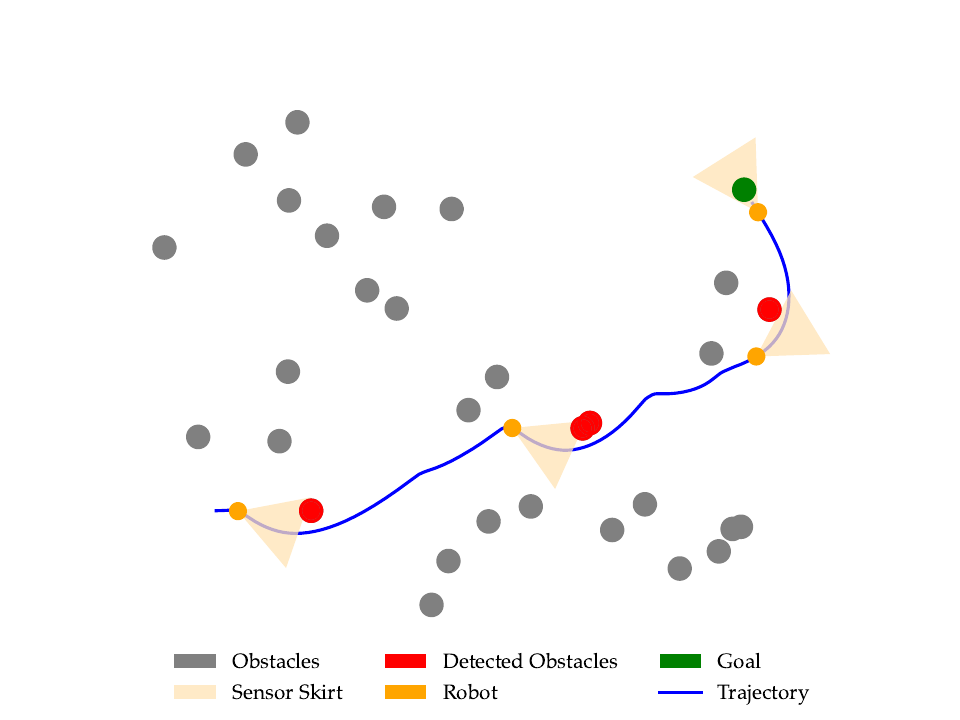}
    \caption{\small Schematic figure for the unicycle navigation scenario with partial observations. The bright orange circle represents the unicycle. The gray circles represent the environmental obstacles, and the green represents the goal. The light orange cone represents the observation space of the agent, which is parameterized by the radius of the cone ($r_{obs}$) and the angle of the cone ($\theta_{obs}$). Each obstacle lights up red when the agent observes the obstacle.}
    \label{fig:unicycle-schematic}
    \vspace{-0.7cm}
\end{figure}

We assume that the unicycle agent has a conical observation space parameterized by two observation parameters: $r_{obs}$, the radius of the cone, and $\theta_{obs}$, the angle of the cone. Furthermore, we assume that the agent detects an obstacle with a probability $p_{obs}$. A schematic of this setting has been provided in the Figure~\ref{fig:unicycle-schematic}

First, we consider the problem of estimating observation space parameters of the unicycle agent, $\theta = (r_{obs},\theta_{obs})$. 
We consider four different scenarios with different $(r_{obs},\theta_{obs})$ values and generate four datasets of
500 trajectories each. 
During each trajectory generation, we randomly choose the location of the obstacles. We assume that the agent follows a known deterministic policy that is attractor-repulsor in nature, i.e., a policy that guides the agent to move toward the goal while avoiding obstacles. The visualization of one such trajectory is provided in Figure~\ref{fig:unicycle-schematic}. We use the cross-entropy method to solve \eqref{eq:negative-log} and estimate the actual parameters $(r_{obs},\theta_{obs})$. For each set of parameters, we provide the ground truth values and the estimated values in Table~\ref{tab:obs}. As it can be seen, our method can closely estimate the actual parameter values.

\begin{table}[!h]
    \centering
    \begin{tabular}{|c|c|c|}
\hline
\textbf{Scenario}&\textbf{Real} $(r_{obs},\theta_{obs})$ & \textbf{Estimated} $(r^*_{obs},\theta^*_{obs})$ \\
\hline
 1&   (55, 0.392) & (57.037, 0.402) \\
\hline
2&(99, 0.785) & (100, 0.826) \\
\hline
3&(22, 0.523) & (23.063, 0.523) \\
\hline
4&(70, 0.262) & (71.146, 0.267) \\
\hline
\end{tabular}
    \caption{\small Actual and estimated values of the observation space parameters.}
    \label{tab:obs}
    \vspace{-0.3cm}
\end{table}

We further examine the trajectories that these estimated $\theta^* = (r^*_{obs},\theta^*_{obs})$ induce to examine how close they are to the demonstration data. 
We compute the Fréchet distance between the actual and predicted trajectories to investigate their similarity. The Fréchet distance is a good metric for comparing curves because it captures the overall shape and continuity by considering the closest possible continuous matching between the \maulik{two curves~\cite{eiter1994computing}}. \maulik{However, the absolute value of the Fréchet distance can be difficult to interpret in isolation, so we normalize it by the straight-line distance between the start and goal locations to obtain a scale-invariant measure of trajectory similarity.}

\begin{table}[ht]
\centering
\begin{tabular}{|c|c|}
\hline
\textbf{Scenario} & \textbf{Normalized Fréchet Distance (Mean $\pm$ Std)} \\ \hline
1        & $2.80 \pm 6.12$                           \\ \hline
2        & $5.93 \pm 13.67$                          \\ \hline
3        & $1.29 \pm 3.54$                           \\ \hline
4        & $1.43 \pm 3.96$                              \\ \hline
\end{tabular}
\caption{\small Average normalized Fréchet distance between 500 actual and predicted trajectories for scenarios in Table~\ref{tab:obs}.}
\vspace{-0.4cm}
\label{tab:frechet-obs}
\end{table}

\maulik{The normalized average Fréchet distances} between actual and predicted trajectories for all scenarios are provided in Table~\ref{tab:frechet-obs}. 
As can be seen, for all four datasets, the estimated parameters are within about $6\%$ of the interval to the original trajectories. These results confirm that by reasoning about the observation space and perception capabilities of the agent, we can mimic the agent behavior and generate agent trajectories that are close to the demonstrated trajectories.

Next, we estimate the probability of observing the obstacles, $p_{obs}$. For this method, we consider the same policy as before, and we further assume access to the actual observation space parameters $(r_{obs},\theta_{obs})$. Same as before, we consider four different scenarios which correspond to four different values of $p_{obs}$ to generate four datasets 
of 500 trajectories each. Same as before, we randomly choose the locations of obstacles during each trajectory. We employ an open-source Python toolbox for Bayesian optimization~\cite{baysianopt} to solve \eqref{eq:negative-log-2} and estimate the $p_{obs}$ values. The actual values of $p_{obs}$ and their corresponding estimated values are presented in Table~\ref{tab:prob}. Our method can predict the $p_{obs}$ values within about $10\%$ of the actual values.
\begin{table}[!h]
    \centering
    \begin{tabular}{|c|c|c|}
    \hline
        \textbf{Scenario} & \textbf{Actual} ($p_{obs}$) & \textbf{Estimated} ($p^*_{obs}$) \\
        \hline
        1 & 0.2 & 0.2209 \\
        \hline
        2 & 0.4 & 0.4352\\
        \hline
        3 & 0.6 & 0.6140\\
        \hline
        4 & 0.8 & 0.8294 \\
        \hline
    \end{tabular}
    \caption{\small Actual and estimated values of observation probability $p_{obs}$.}
    \label{tab:prob}
    \vspace{-0.5cm}
\end{table}

We further examine the trajectories that these estimated $p_{obs}$ induce to examine how close they are to the demonstration data. 
The Fréchet distance between actual and predicted trajectories is provided in Table~\ref{tab:frechet-prob}. 
As can be seen, for all the datasets, the predicted trajectories using estimated probability values are within $5\%$ of distance to the original trajectories, which confirms that our method learns a close and predictive value of $p_{obs}$

\begin{table}[h!]
\centering
\begin{tabular}{|c|c|}
\hline
\textbf{Scenario} & \textbf{Normalized Fréchet Distance (Mean $\pm$ Std)} \\ \hline
1        & $2.88 \pm 3.21$                       \\ \hline
2        & $4.95 \pm 6.92$                       \\ \hline
3        & $3.01 \pm 6.20$                       \\ \hline
4        & $2.55 \pm 7.36$                       \\ \hline
\end{tabular}
\caption{\small Average normalized Fréchet distance between 500 actual and predicted trajectories for scenarios in Table~\ref{tab:prob}.}
\label{tab:frechet-prob}
\vspace{-0.5cm}
\end{table}

Finally, we focus on estimating the agent's policy. We generate a dataset of 500 trajectories with different obstacle configurations for all trajectories using the aforementioned attractor-repulsor policy with fixed values of $r_{obs}, \theta_{obs}$, and $p_{obs}$. We aim to learn a policy that best imitates the dataset. For this purpose, we employ Diffusion policies where we learn an action diffusion policy that takes action based on observation and the agent's current state. \maulik{Note that, unlike the previous components where we consider multiple scenarios to estimate different parameters, here, we focus on learning a single policy from the dataset.} 
However, this can easily be extended to different policies. Using the learned diffusion policy, we induce a dataset of predicted trajectories. As before, we compute the normalized Fréchet distances between the actual and predicted trajectories. The average normalized Fréchet distance is $3.47 \pm 9.31$, confirming that the resulting trajectories resemble demonstration trajectories.

\textcolor{black}{To further assess the effectiveness and robustness of our modular approach, we evaluate the generalization capability of the learned diffusion policy. Specifically, we train our diffusion policy on a set of observation space parameters and test its performance on previously unseen observation parameters. We then compare the trajectories generated by the learned policy with those produced by the original policy. The results, summarized in Table~\ref{tab:modularity}, show the average normalized Fréchet distances between the predicted and original trajectories across different observation parameters. Despite variations in the observation space, the learned policy successfully generalizes to new settings, producing trajectories that closely resemble those from the original policy. This demonstrates the adaptability of our modular approach and showcases its ability to capture underlying behavioral patterns while maintaining robust performance across diverse conditions.}

\begin{table}[h!]
\centering
\begin{tabular}{|c|c|}
\hline
\textbf{Test ($r_{obs},\theta_{obs}$)} & \textbf{Normalized Fréchet Distance (Mean $\pm$ Std)} \\ \hline
(80,0.523)        & $7.65 \pm 15.66$                       \\ \hline
(70,0.785)       & $9.79 \pm 20.50$                       \\ \hline
(60,0.628)        & $8.77 \pm 16.80$                       \\ \hline
\end{tabular}
\caption{\small Average normalized Fréchet distance between 100 different trajectories generated by the trained diffusion model against the original policy for different observation parameter values.}
\label{tab:modularity}
\vspace{-0.5cm}
\end{table}

\section{Imitating Human Behavior}\label{sec:expt}
\begin{figure}
    \includegraphics[width=1\linewidth]{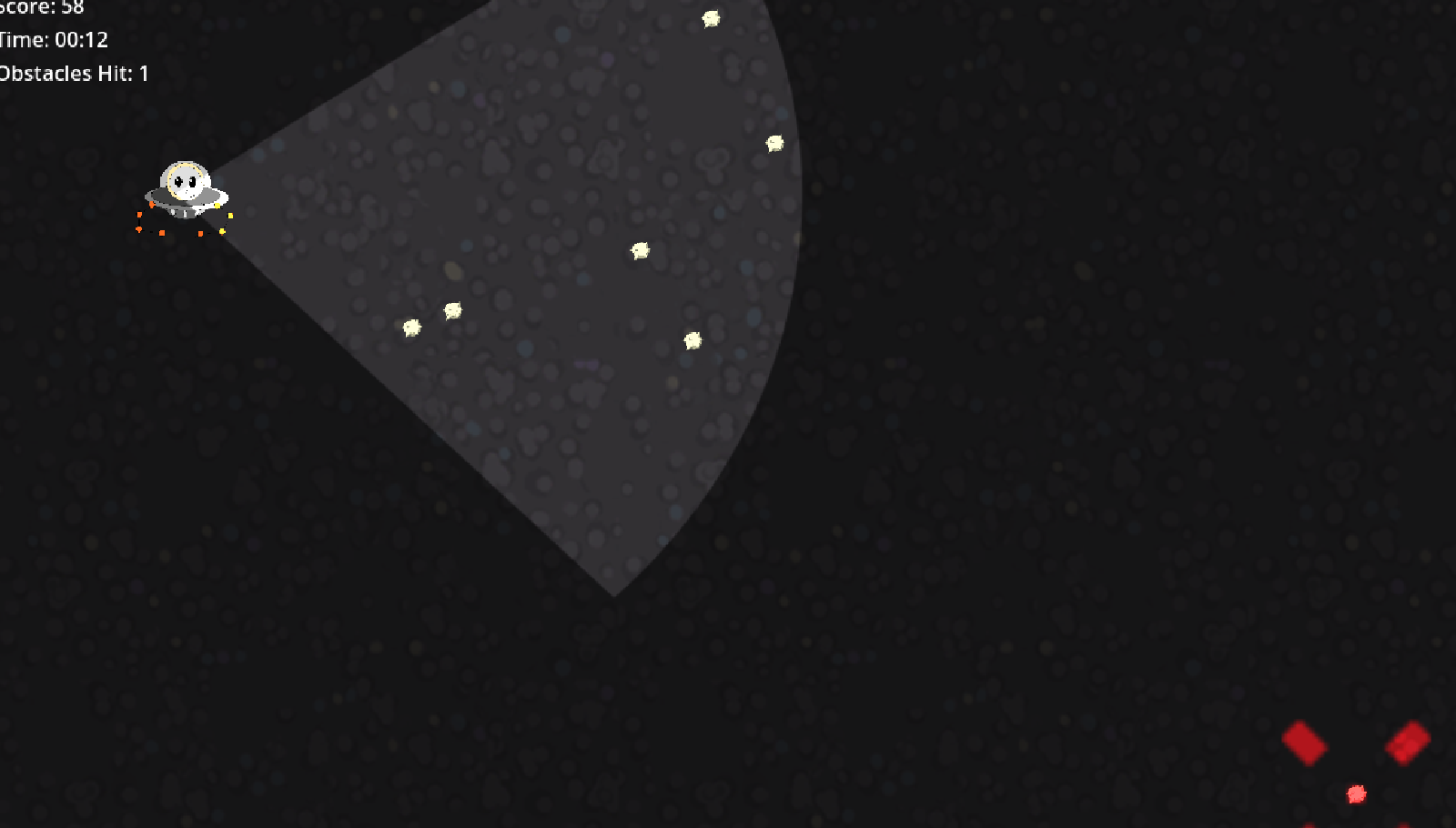}
    \caption{\small Participants used a mouse and keyboard to navigate to reach the goal in the bottom right corner. Their vision was kept limited to simulate human perception in real-world scenarios. The obstacles 
    followed a unique movement pattern unknown to the participants. Some obstacles moved quickly, others slowly, and some remained stationary. We collected trajectory data from the participants to better understand how humans observe their surroundings and make decisions in this driving-like scenario.}
    \label{fig:godot-game-schematic}
    \vspace{-0.6cm}
\end{figure}
\honghao{This section focuses on replicating human behavior using data collected directly from human demonstrations.}
We collect human trajectory data to capture realistic navigation strategies, aiming to predict and understand human actions for improved collaboration. By modeling these behaviors, our approach enables the system to anticipate and adapt to human decision-making, enhancing its ability to operate effectively in dynamic and complex environments.


\subsection{Godot Game Environment with Data Collection}
To facilitate data collection from human participants, we developed a custom game platform using the Godot game engine in version 4.3 \cite{godotengine}. A schematic of the game is shown in Fig.~\ref{fig:godot-game-schematic}. In this game, volunteers controlled a point-mass robot to navigate the environment, avoid obstacles, and reach a designated goal area.

\subsubsection{Environment Design and Game Mechanics}
The game environment consisted of a rectangular area with a fixed starting point for the robot and a fixed goal location. Participants controlled the robot using standard keyboard inputs, enabling them to move in any direction. \honghao{The participants used the mouse to control the robot's viewing direction $\psi$, which determined how the field of view rotated within the environment.}
The robot’s speed was constrained to mimic realistic movement dynamics, adding further complexity to the task. Participants had to carefully balance speed with accuracy to avoid collisions while reaching the goal efficiently. \honghao{We aim to learn both navigation and camera control strategies from this setup using the collected data to replicate how humans coordinate movement and visual attention during the task.}

\subsubsection{Limited Field of View}
To simulate real-world constraints, participants' vision was restricted to a cone-shaped field of view. This limited their ability to observe the entire environment at once, as only obstacles within the cone were visible. However, the goal remained visible at all times, giving participants a consistent reference point to navigate toward. The field of view had fixed $\theta_{\text{obs}}$ and fixed $r_{\text{obs}}$. 

\vspace{-0.0cm}

\subsubsection{Data Collection}
\honghao{Participants played the game on their computer and gave informed consent under the oversight of an Institutional Review Board.\footnote{Protocol 65022. approved by the Administrative Panels for the Protection of Human Subjects at Stanford University, On February 29, 2024.}}
During the experiment, we recorded two critical types of trajectory data: Agent Trajectory Data: This included the position (x,
y) and orientation ($\psi$) of the human-controlled robot
at each timestep. This data captured the full path taken
by participants as they navigated toward the goal. Obstacle Trajectory Data: Both static and moving
obstacles had their positions recorded throughout each
trial. Static obstacles maintained constant positions,
while dynamic obstacles followed specific motion patterns, which were also logged for later analysis. We used data from 8 participants to ensure a diverse set of navigation behaviors.

\begin{figure}
    \centering
    \begin{subfigure}[b]{0.95\linewidth}
        \centering
        \includegraphics[width=\linewidth]{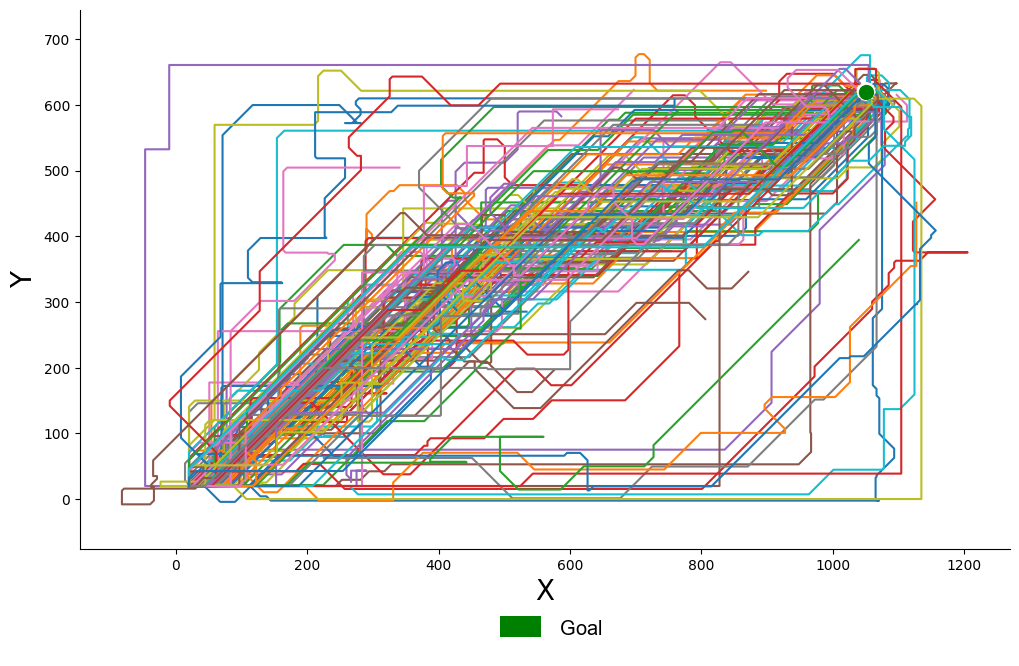}
        \vspace{-0.7cm}
        \caption{Trajectories from human data (training)}
        \label{fig:frechet-human-human}
    \end{subfigure}
    \hfill
    \begin{subfigure}[b]{0.95\linewidth}
        \centering
        \includegraphics[width=\linewidth]{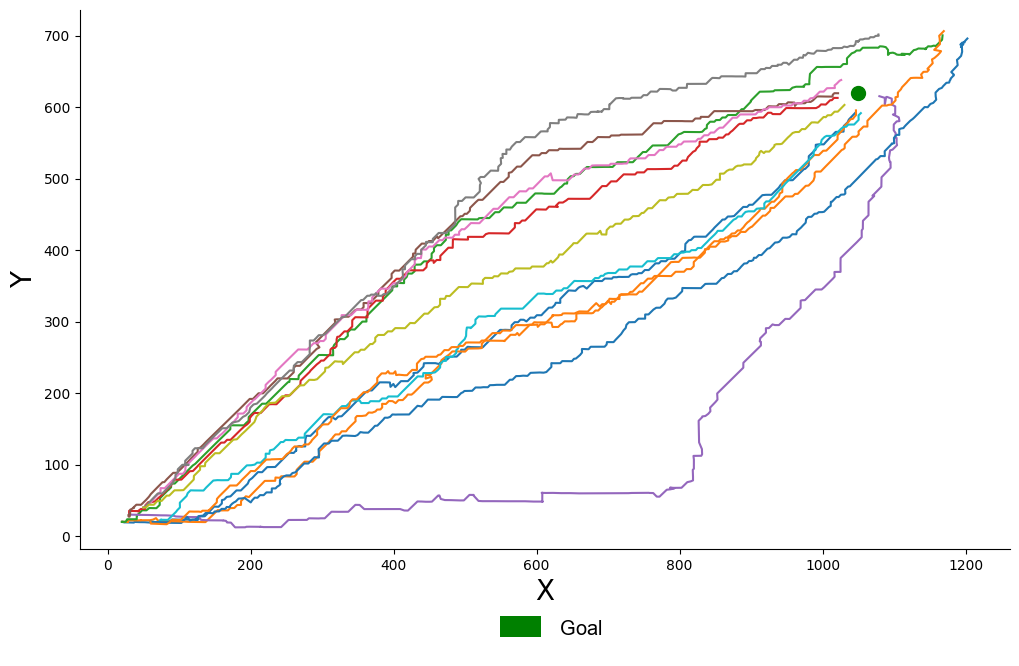}
        \vspace{-0.7cm}
        \caption{Trajectories from inference (testing)}
        \label{fig:frechet-human-unet}
    \end{subfigure}
    \caption{\small Qualitative illustration of the human data and model inference. All the trajectories have the same environment within this figure. }
    \label{fig:Trajectories compare}
    \vspace{-0.7cm}
\end{figure}

\vspace{-0.3cm}
\subsection{Transformer-Based Diffusion for Navigation}
The collected human trajectory data serves as the foundation for training our navigation model, aimed at replicating human-like obstacle avoidance and goal-reaching strategies. We employ a transformer-based diffusion model that uses a multi-head cross-attention mechanism, inspired by the Transformer architecture \cite{vaswani2017attention}, with a class-conditional U-Net \cite{Ronneberger2015unet}. During testing, noisy action inputs and observation embeddings are processed through each U-Net layer. The cross-attention mechanism allows the model to focus on various aspects of the environment, such as obstacle proximity or goal direction, dynamically adapting to the most relevant features at each timestep. The model outputs the action changes, represented as $\Delta x, \Delta y, \Delta \psi$.

\vspace{-0.1cm}
\subsection{Trajectory Generation and Validation}
The experiment is designed in three parts to validate the model's ability to replicate human-like obstacle avoidance and goal-reaching behavior:

\subsubsection{Original game environment}
We computed the Fréchet distances between the actual and predicted trajectories (Figure~\ref{fig:Trajectories compare}).
The average normalized Fréchet distance was found to be 11.3 ± 0.06, indicating a strong alignment between the model’s trajectory predictions and human trajectories in the original game environment. Additionally, the minimum Fréchet distance observed was 4.48, demonstrating the model’s capability to closely replicate human-like navigation under certain conditions.

\subsubsection{Scaled-Down Lab Environment}
\begin{figure}
    \centering    \includegraphics[width=\linewidth]{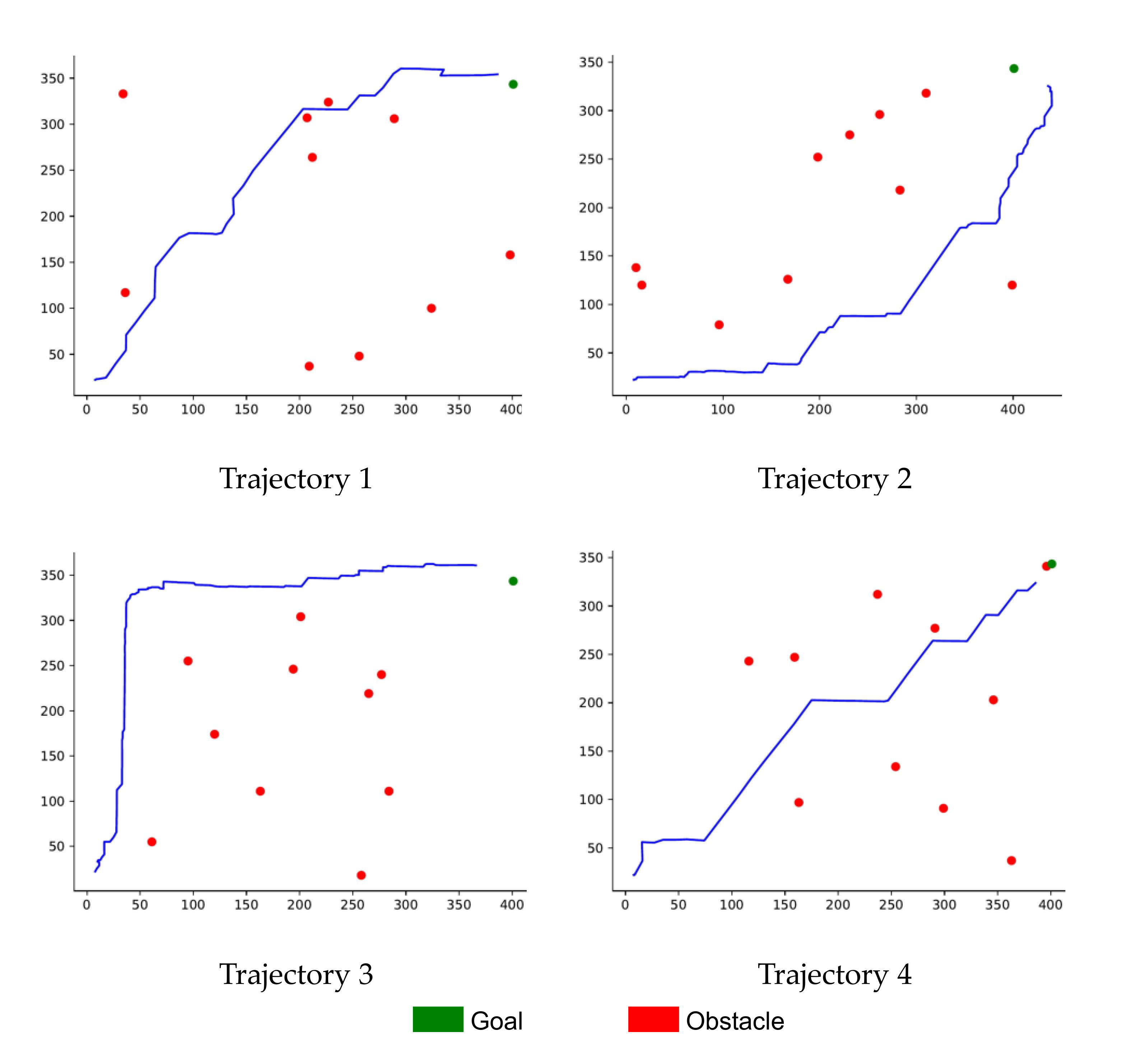}
    \vspace{-0.7cm}
    \caption{\small Trajectory generation in an unknown environment.}
    \label{fig:scaled-down_random}
    \vspace{-0.5cm}
\end{figure}

To demonstrate the agent's ability to generalize beyond the training environment, we generated four scenarios with randomly positioned obstacles (See Fig.~\ref{fig:scaled-down_random}). In each scenario, the agent begins at the bottom-left corner and aims to reach the goal located at the top-right corner. These examples illustrate that the agent can adapt its trajectory to avoid obstacles while efficiently navigating toward the goal, dynamically controlling its observation space to prioritize obstacle avoidance in a manner that closely mimics human behavior. This behavior showcases the robustness of the policy in handling previously unseen environments.

\subsection{Safety Analysis of Trajectories Data}
\begin{figure}
    \includegraphics[width=1.0\linewidth]{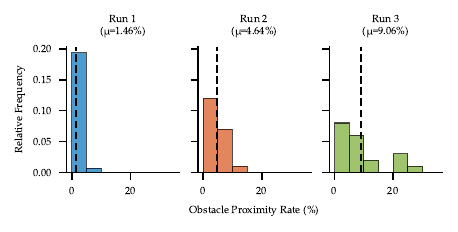}
    \vspace{-0.8cm}
    \caption{\small \honghao{Comparison of obstacle proximity rates across human and model-generated trajectories. Obstacle proximity rate is defined as the fraction of time steps during which the agent was within 20 units of at least one obstacle. The value of $\mu$ shown above each subplot represents the mean obstacle proximity rate for all trajectories in the corresponding run.} Run 1: 150 Human demonstration trajectories. Run 2: 20 trajectories generated by the diffusion policy with the observation model. Run 3: 20 trajectories generated by the diffusion policy without the observation model.  
    All obstacles used in these experiments follow the same movement patterns as those from the human data collection environment. \honghao{Generated trajectories with the observation model better replicate human navigation behavior, with a lower risk of entering danger zones.}}
    
    \label{fig:histogram_baseline}
    \vspace{-0.6cm}
\end{figure}

\honghao{In this section, we evaluate whether the observation model helps the learned policy replicate the safety-aware behavior that was demonstrated by humans.}
To evaluate the safety of our navigation strategies, we analyzed three distinct sets of trajectories: \emph{Human Data}: Data collected from human participants. \emph{Diffusion Model with Observation model}: This refers to trajectories generated by the diffusion model while incorporating an observation model learned from human demonstration data. The observation model captures how humans tend to direct their attention or gaze during navigation. 
\emph{Diffusion Model without Observation}: In this baseline, the diffusion model does not include the human learned observation behavior. Instead of predicting $\Delta \psi$, it applies a fixed angular velocity, \honghao{independent of the environment}. We measured the percentage of time during which the robot was within a critical 20-unit distance from any obstacle.
The results are summarized in Fig.~\ref{fig:histogram_baseline}. The human data shows a critical proximity percentage of 1.46$\%$, while \emph{Diffusion Model with Observation model} recorded 4.64$\%$. In contrast, \emph{Diffusion Model without Observation} as the baseline only reached 9.06$\%$. These results suggest that incorporating human-like observation strategies, which enhance real-time situational awareness, significantly improve obstacle avoidance and overall trajectory safety. 

\section{Hardware Experiment}
To showcase the capabilities of our method in real-time robotic settings, we perform a hardware experiment with a car. We considered the setting of a car navigating through an environment with static obstacles. We aim to replicate the human behavior collected from data in Section~\ref{sec:expt}.

We use bicycle dynamics to model the car:
\begin{align}\label{bicycle}
    p_{t+1} & = p_t + h\cdot v_k\cos{\phi^i_t}, \;
    q_{t+1} = q_t + h\cdot v_t\sin{\phi^i_t} \nonumber \\
    \phi_{t+1} & = \phi + \frac{v_t}{L}\cdot\tan(\delta_t), \;
    \delta_{t+1} = \delta_{t} + h\cdot\omega_{t} \nonumber \\
    \psi_{t+1} & = \psi_{t+1} + \Delta \psi_t,
\end{align}
where $p_t$ and $q_t$ are $x$ and $y$ coordinates of the positions in the 2D plane, $\phi_t$ is the heading angle from positive x-axis, $v_t$ is the forward velocity, $\delta_t$ is the steering angle, $\psi_t$ is the orientation of the sensor skirt, $\Delta \psi_t$ is the change in sensor skirt orientation, $\omega_t$ is the steering rate of the agent at time $t$, and $L$ is the lenght of the car.  We employ the transformer diffusion model from Section~\ref{sec:expt} for the bicycle dynamics. Note that in this case, the output from the trained diffusion policy is $\Delta x, \Delta y, \Delta \psi$. \maulik{However, since the bicycle model requires linear and angular velocities, we convert the diffusion model outputs ($\Delta x, \Delta y, \Delta \psi$) into the corresponding velocity commands via}
\begin{align}
    v_t = K_v\cdot||\delta x ||, \; &  
    \phi_d = \tan^{-1}(\Delta x_2, \Delta x_1) \nonumber \\
    \omega_t &= K_\omega\cdot(\phi_d - \phi_t),
\end{align}
\vspace{-0.4cm}

\noindent where $||\cdot||$ represents the Euclidean norm, $\Delta x_{i}$ is the $i^{th}$ component of the vector $\Delta x$, $\phi_d$ is the desired heading angle for the robot to move towards the next waypoint, and $K_v$ and $K_\omega$ are hyperparameters tuned for the car. Furthermore, to make the policy inference fast enough for real-time implementation, we use a Denoising Diffusion Implicit Model (DDIM) \cite{song2020denoising} based approach to speed up the denoising process. Instead of 100 DDPM steps, we use 10 DDIM steps followed by 10 DDPM steps to obtain control inputs for the car. This approach makes the policy applicable to the car in real-time. We use the VICON motion capture system to obtain real-time position updates for the car. We use the Robotics Operating System~\cite{quigley2009ros} to communicate with the car and send commands.

We employ the trained policy from humans on the car. The experiment results can be found in Figure~\ref{fig:hardware-experiment}. As can be seen, the car successfully navigates the environment with obstacles. Like humans, the car changes its observation direction to look around, detect obstacles, and avoid them while moving toward its goal in real-time.









\section{CONCLUSIONS}

In this work, we show that for agents who have limited perception capabilities characterized by a limited field of view, viewing distance, and failure to detect objects within their field of view, it is important to model these observation limitations in addition to their policy in order to correctly predict the motion and behavior of the agent. We demonstrated the utility of approximating these observation parameters for robots, and we showed that by observing humans navigating a cluttered scene \maulik{and explicitly reasoning about their limited observation space and observation strategies,} we could better imitate their behavior. We further showed that the resulting policy could be executed on physical robotic hardware.










\bibliography{bibliography,monroe_bib}
\bibliographystyle{IEEEtran}

\end{document}